\pdfoutput=1
\documentclass[11pt]{article}

\usepackage{ACL}

\usepackage{times}
\usepackage{latexsym}

\usepackage[T1]{fontenc}
\usepackage[utf8]{inputenc}

\usepackage{microtype}

\usepackage{inconsolata}

\usepackage{amsmath}
\usepackage{amssymb}
\usepackage{interval}
\usepackage{xcolor}
\usepackage{subcaption}

\usepackage{multirow}
\usepackage{enumitem}
\usepackage{graphicx}
\usepackage{wrapfig}
\usepackage{bbm}
\usepackage{booktabs}

\usepackage{tabularx,lipsum}
\usepackage{mdframed}

\usepackage{algorithm}
\usepackage{algorithmicx}
\usepackage{algpseudocode}
\usepackage[capitalise]{cleveref}
\graphicspath{{./}{pics/}}
\usepackage{etoolbox}

\AtBeginEnvironment{quote}{\small\itshape}

\renewenvironment{quote}
               {\list{}{%
                    \leftmargin0pt       % 为了去掉左边距
                    \rightmargin0pt      % 为了去掉右边距
               }
               \item\relax}
               {\endlist}

\newcommand{\header}[1]{\smallskip\noindent\textbf{#1}}
\newcommand{\bullethdr}[1]{\smallskip\noindent\textbullet\,\emph{#1}}

\title{The Knowledge Alignment Problem: Bridging Human and External Knowledge for Large Language Models}
\author {
  Shuo Zhang\textsuperscript{1}\footnotemark[1]
  \quad
  Liangming Pan\textsuperscript{\rm 2}\quad
  Junzhou Zhao\textsuperscript{\rm 1}\footnotemark[2]
  \quad
  William Yang Wang\textsuperscript{\rm 2}
  \\ \textsuperscript{\rm 1}MoE KLINNS Lab, Xi'an Jiaotong University, P. R. China \\
  \textsuperscript{\rm 2}University of California, Santa Barbara, USA\\
  \texttt{\{zs412082986@stu, junzhou.zhao@mail\}.xjtu.edu.cn}\\ \texttt{\{liangmingpan, wangwilliamyang\}@ucsb.edu}
}
\begin{document}
\maketitle
\begin{abstract}
  Large language models often necessitate grounding on external knowledge to generate faithful and reliable answers. 
  Yet even with the correct groundings in the reference, they can ignore them and rely on wrong groundings or their inherent biases to hallucinate when users, being largely unaware of the specifics of the stored information, pose questions that might not directly correlate with the retrieved groundings.
  In this work, we formulate this knowledge alignment problem and introduce \textsc{MixAlign}, a framework that interacts with both the human user and the knowledge base to obtain and integrate clarifications on how the user question relates to the stored information. \textsc{MixAlign} employs a language model to achieve automatic knowledge alignment and, if necessary, further enhances this alignment through human user clarifications. Experimental results highlight the crucial role of knowledge alignment in boosting model performance and mitigating hallucination, with improvements noted up to 22.2\% and 27.1\% respectively.
  We also demonstrate the effectiveness of \textsc{MixAlign} in improving knowledge alignment by producing high-quality, user-centered clarifications.\footnotemark[3]\footnotetext[1]{The work was done during a visit to UCSB.}\footnotetext[2]{Corresponding author}\footnotetext[3]{Code and data available at \url{https://github.com/ShuoZhangXJTU/MixAlign}}
\end{abstract}

\section{Introduction}
Despite the recent advances of large language models (LLMs), they still struggle in unfamiliar scenarios not covered during pre-training~\cite{bubeck2023sparks}.
A common approach to mitigate this issue involves retrieving and incorporating supporting evidence from an external knowledge base~\cite{guu2020retrieval,shuster2021retrieval}.
% This approach grounds the generation process with supporting evidence from a grounding-specific knowledge base.
While the method indeed often improves the end-task performance, it still suffers from  issues such as generating text
that includes extraneous information not present in the retrieved
knowledge~\cite{dziri2022origin}, ignoring the knowledge
entirely~\cite{krishna-etal-2021-hurdles}, or even contradicting the
knowledge~\cite{longpre2021entity,wu2023ragtruth}.
These erroneous behaviors are interpreted as a passive defense mechanism against
poor retrievals~\cite{gao2022rarr}.

\begin{figure}[t]
  \includegraphics[width=\linewidth]{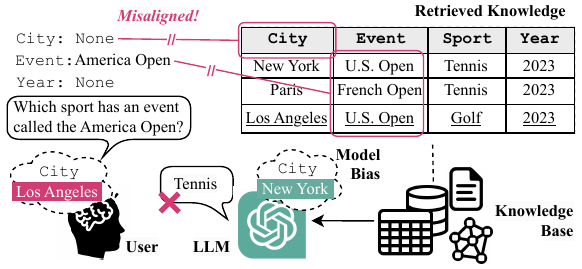}
  \centering
  \caption{Knowledge Misalignment. Even if the user knows about the city constraint, he/she may not put it in the question being unaware that ``city'' is needed to filter noisy candidates. For the same reason, the user may not give a precise event name. Due to the misalignment between the user question and the grounding knowledge, the LLM fails to correlate the question with the correct grounding (underlined) and relies on its own biased knowledge (presuming New York as the intended city reference) to generate an incorrect answer.}
  \label{fig:intro}
\end{figure}

In this work, we argue that the primary cause of the error cases stems from the misalignment between human and grounding knowledge.
This misalignment is quite common, as users are often unfamiliar with the information contained in the external database. When framing their questions, they might unintentionally phrase them in ways that either inconsistently state or even overlook the conditions and constraints of the retrieved groundings (refer to~\cref{fig:intro}). 
Facing this, the language model may follow
spurious correlations and incorporate biased model knowledge to generate biased, misleading, or unsupported content.

To address this issue, we study the \textit{Knowledge Alignment} problem considering various alignment types as depicted in~\cref{tab: type}.
Unlike recent works on value alignment which aim to ensure LLM generation follows human values, ethics, and goals~\cite{ouyang2022training,pyatkin2022reinforced}, knowledge alignment seeks to bridge human and grounding knowledge for LLM, thereby enhancing its ability to utilize grounding knowledge for faithful decision-making and issue resolution.

Towards solving the knowledge alignment problem, we propose \textsc{MixAlign}, a framework that interacts with both the user and the knowledge base to acquire
clarifications on how the user's question relates to the stored grounding knowledge.
% As shown in~\cref{fig:intro}, 
\textsc{MixAlign} initiates the process with \emph{model-based knowledge alignment}, where the LLM is employed to map the conditions and constraints of the user question to corresponding ones within the knowledge base.
In cases the mapping process yields uncertainties or the evidence remains unclear, it generates a question seeking further clarification from the user, a step we refer to as \emph{human-assisted knowledge alignment}.
The clarifications from these steps are incorporated to generate the final answer.

In summary, our major contributions are:
\begin{itemize}
\item We study the Knowledge Alignment problem, a prevalent yet critical issue that influences the efficacy of LLMs when interacting with external databases.
\item We introduce \textsc{MixAlign}, a mixed-initiative clarifying framework designed to improve knowledge alignment.
\item Comprehensive evaluations highlight the importance of knowledge alignment and demonstrate the effectiveness of \textsc{MixAlign} in generating high-quality clarifications.
\end{itemize}

% When the LM is familiar with the topic, it tends to override grounding and use
% its own saved knowledge.
% When unfamiliar, and the question is simple without detailed constraints, it
% tends to use the grounding info without verifying it.
% Such a familiar can be measured by lexical overlap~\cite{shi2023large}.
% When constraints are specified in the question, the model (advanced model) tends
% to decide if the constraint is not satisfied to reject fatal grounding info.

% Knowledge facts are stored in the model and the outside.To align with the model,
% we may ask the model to paraphrase the question so that it reaches the same
% answer.To align with the database, we may transfer it to the execution query.
% By comparing the model results, we may know if the model is familiar with the
% topic.

% Now more case is that the model does not hallucinate with its own biased
% dataset, it only introduces extra facts

% With a simple question, we need to clarify it to trigger both.
% If triggered DB and model, then fact-checking and verify.
% If triggers model only, shall decide if it is not in the DB, if not in .

%%% Local Variables:
%%% mode: latex
%%% TeX-master: "emnlp2023"
%%% End:

\section{Related Work}

\header{Alignment in Large Language Models.}
Recent efforts have been made to ensure that AI systems pursue goals that match human values or interests rather than unintended and undesirable goals~\cite{ngo2022alignment,wolf2023fundamental}.
This issue, known as the alignment problem in LLMs, has been addressed in several ways.
Reinforcement Learning from Human Feedback (RLHF) is one such approach, which fine-tunes the LLM according to the reward signals adhering to human evaluators' preferences~\cite{ouyang2022training,bai2022training}.
Another strategy involves in-context learning using textual prompts that are helpful, honest, and harmless~\cite{askell2021general,rae2021scaling}. 
The development of interpretability techniques to scrutinize the concepts learned by networks is yet another crucial approach, with the long-term aim of detecting and rectifying misaligned goals prior to deployment~\cite{meng2022locating,burns2022discovering}.
This work can be seen as a special case of interpretability methods. Unlike existing works that emphasize aligning human values with LLM behavior, we aim to align human knowledge with external domain knowledge. This knowledge alignment enhances semantic and logical consistency between human expression and the stored evidence, thereby enabling LLMs to engage in more effective reasoning and problem-solving.

\header{Clarification Question Generation.}
The study of asking clarifying questions spans a wide range of tasks, including
information retrieval and open-domain question
answering~\cite{pan2019recent,majumder2021ask,
  kuhn2022clam,pyatkin2022reinforced,meng2023followupqg}.
The effectiveness of these questions is often determined by information-theoretic
measures such as relevance, informativeness, or
utility~\cite{rao2018learning,rao2019answer,white2021open}.
Rule-based methods have been proposed for generating clarification questions by
filling manually defined templates~\cite{wang2021template} or applying a set of
syntactic transformations on ambiguous questions~\cite{dhole2020resolving}.
In addition to rule-based methods, neural network-based approaches have been
proposed to generate more coherent questions by training text generation
models~\cite{rao2018learning,pan2020semantic} or utilizing state-of-the-art
pre-trained large language
models~\cite{krasheninnikov2022assistance,kuhn2022clam}.
Most of the existing works focus on resolving ambiguities within user queries,
whereas we seek clarifications on how the user question is related to the stored
knowledge.
Instead of requesting the user to provide more context aimlessly, we direct them
on how to offer such information by concentrating on a particular constraint.

%%% Local Variables:
%%% mode: latex
%%% TeX-master: "emnlp2023"
%%% End:

\begin{table*}[t]
\small
\centering

\label{tab: type}
\begin{tabularx}{\linewidth}{p{1.5cm}p{5.3cm}Xp{1.5cm}}
\toprule
Type & Explanation & Example & Percentage\%\\
\midrule 
Semantic & The user might use an ambiguous term that, while ideally should map to a single item in the database, in reality can correspond to multiple attributes or values.& For ``What is the best burger?'', when you say ``best burger'', are you referring to taste, nutritional value, price, or a combination of these attributes? & 41.48\\
\midrule
Contextual & The user may have implicitly established some conditions without explicitly expressing them. & For ``What is the 15th most populous city in the United States?'', the statistics may vary with time, which year are you considering? & 32.04\\
\midrule
Structural & The user might have stated some conditions that are not addressed in the database. & For ``Find me an American writer.'', the question can not be answered when the database does not include the nationality of the writers. & 56.70\\
\midrule
Logical & The user can ask complex questions where certain conditions need to be determined before other conditions can be clarified, while the database only supports basic logits such as ``and'', ``or'' and ``not''. & For ``Fine me for a movie directed by the singer who has won a Grammy', the question answering requires first identifying singer who have won a Grammy, and then finding the films directed by the identified singers.  & 5.57\\

% relationships between conditions, as conveyed in the user's expression, may be unclear or excessively complex, e.g., nested questions.
\bottomrule
\end{tabularx}
\caption{Knowledge misalignment types. We evaluate 1,173 valid examples from our FuzzyQA dataset with an evaluation protocol based on GPT-4. 
The overall proportion of samples with knowledge misalignment is 79.54\%. ``Percentage'' denotes the ratio of examples with a certain type to those with any misalignment.}
% }
\end{table*}

\begin{figure}[t]
\centering  
\includegraphics[width=.9\linewidth]{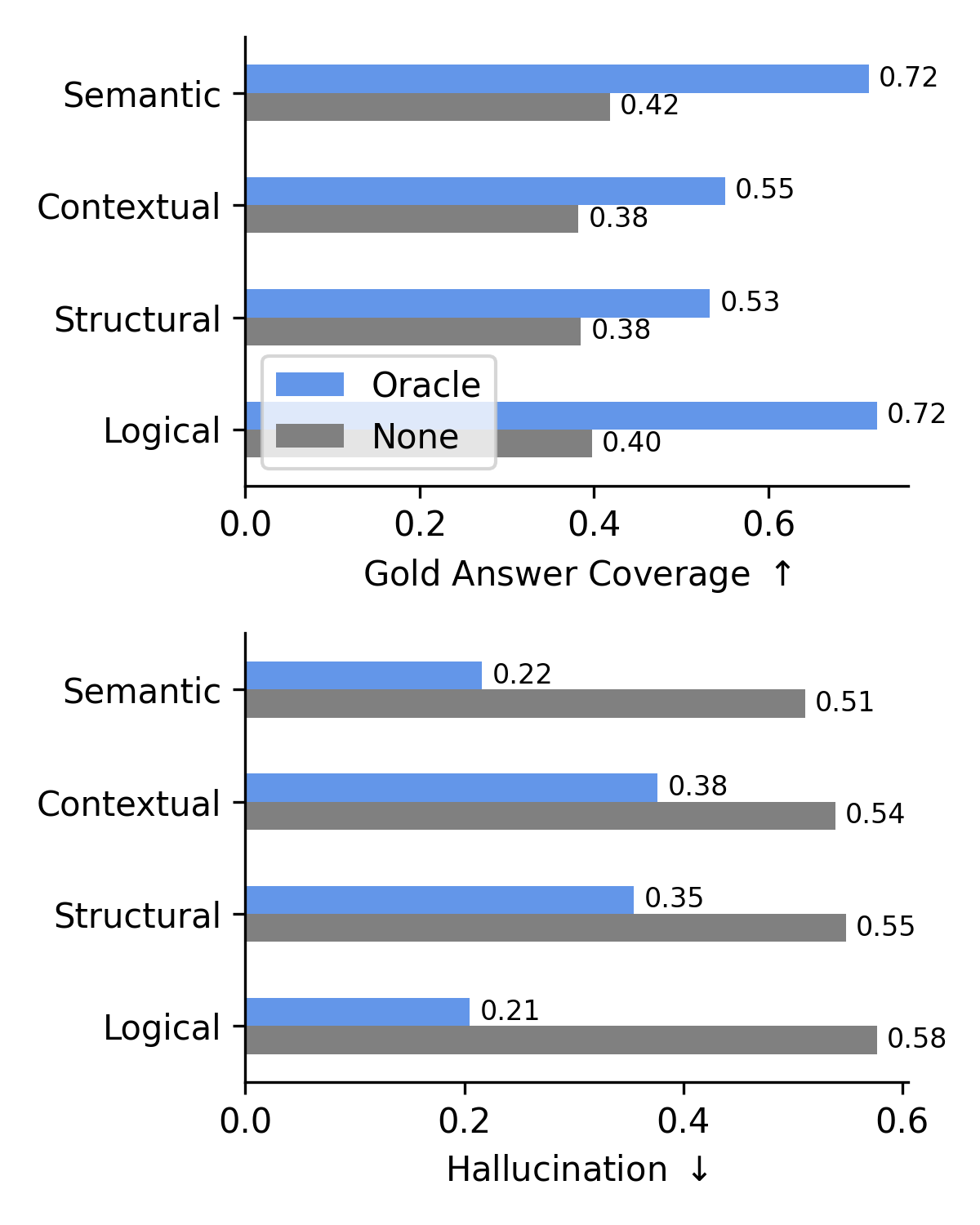}
\caption{Oracle clarification results regarding different knowledge alignment types.}
\label{fig:overall_type}
\end{figure}

\section{The Knowledge Alignment Problem}
We study how aligning human knowledge with grounding knowledge, represented by conditions and constraints found in the user question $(Q)$ and retrieved evidence $(K)$, impacts the LLM's ability to utilize evidence for answering questions.
Specifically, we address knowledge misalignments by acquiring clarifications $(C)$, refining our prompt to: $\text{LLM}(A|Q,K) \rightarrow \text{LLM}(A|Q,K,C)$.

% i.e., $LLM(A|Q,K)$. 
We commence our study with a straightforward setting and consider questions that inquire about a single, specific subject, allowing us to express conditions and constraints in the form of attribute-value pairs related to that subject. For the representation of evidence, we opt for tabular databases due to their inherent clarity and structured nature. Each row in these databases encapsulates well-defined constraints. While other knowledge formats, such as triplets in knowledge graphs or textual paragraphs, can also be organized in this manner, we leave them for future research.

As shown in~\cref{tab: type}, we consider four misalignment types that correspond to conditions with different expressions (semantic), conditions absent in either the user question (contextual) or domain knowledge base (structural), and complex conditions composed of multiple simple conditions (logical), respectively.

To evaluate the importance of knowledge alignment in enhancing language model performance, we conducted experiments using oracle clarifications on different alignment types (refer to Sec.~\ref{sec:exp} for detailed settings).
As depicted in~\cref{fig:overall_type}, we observe a notable difference in performance across different knowledge alignment types, with a marginal gap ranging from $15\%$ to $32\%$ in gold answer coverage and $16\%$ to $37\%$ in hallucination.
% \header{Semantic and logical knowledge alignment types are more commonly encountered.}

Among the alignment types, we find that semantic and logical alignment exhibit a larger performance gap compared to the other two types. 
The semantic and logical alignment share a common characteristic: they prioritize the analysis of existing conditions and constraints rather than requesting the integration of additional, unmentioned information.
This distinction is primarily driven by the nature of our dataset, where the questions are expected to be within the table's information scope. In practice, however, it is common that individuals who are unfamiliar with the domain may require additional contextual information, while those who are familiar with the domain may not.

\begin{figure*}[htp]
\centering  
\includegraphics[width=\linewidth]{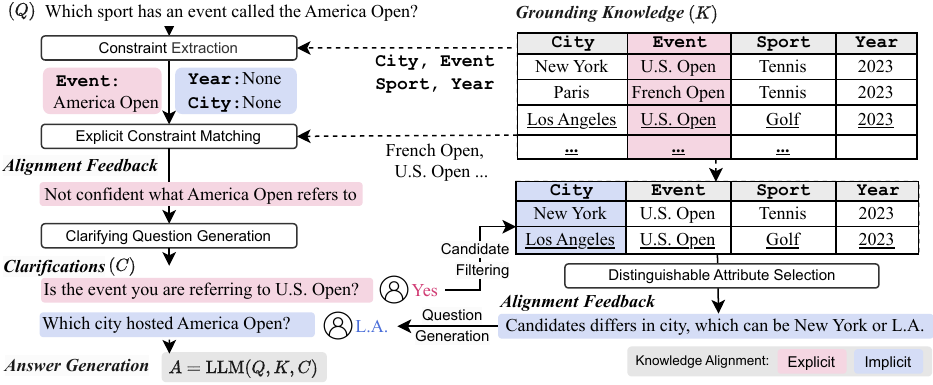}
\caption{Diagram of \textsc{MixAlign}. \textsc{MixAlign} aims to identify knowledge misalignments and obtain clarifications regarding them automatically. It first handles explicit constraints in the user's question for semantic and logical alignment (Explicit Knowledge Alignment), then tackles implicit or missing constraints for contextual and structural alignment (Implicit Knowledge Alignment).
Given the user question, \textsc{MixAlign} utilizes LLM to extract and correlate the constraints within the question referring to the grounding knowledge. If the model cannot confidently establish alignment, a clarifying question is generated to seek assistance from the user. The alignment information is then incorporated to filter candidate knowledge groundings. If confusion persists, the LLM is employed to select an attribute that can distinguish the remaining groundings and seek further clarification from the user.}
\label{fig:MixAlign}
\end{figure*}

% Note that in Model-based Q-K Alignment, the extracted value for an attribute can be a phrase, e.g., \textcolor{violet}{hometown:} \textcolor{blue}{The 15th most populous city in the United States}. In this case, the information extraction step works as a question decomposer and the co-reference resolution step as a subquestion solver, e.g., \textcolor{blue}{The 15th most populous city in the United States} is referring \textcolor{violet}{San Francisco}.
\section{Methodology}
In this section, we introduce \textsc{MixAlign}, a method designed to enhance knowledge alignment in grounded generation. \textsc{MixAlign} utilizes the LLM to align user expressions with the grounding knowledge and, if necessary, further enhances this alignment through human clarification.
\cref{fig:MixAlign} depicts how \textsc{MixAlign} addresses the knowledge format of tabular databases (as detailed in Section 3). For a discussion on its adaptability to other knowledge forms, please refer to Appendix C.

\subsection{Explicit Knowledge Alignment}
In this stage, we align constraints that are explicitly stated in the user question with those from grounding knowledge and obtain clarifications.
As detailed in Section 3, constraints are represented as attribute-value pairs.
Considering the potentially vast number of grounding knowledge constraints, we employ a two-step approach that first extracts values from user questions based on attributes from the grounding knowledge, and then matches values for those valid attributes.
In cases where the model is uncertain about the correlation of a particular constraint, we engage the user by posing a question to confirm the alignment. This interactive step ensures the accuracy and reliability of the alignment process.
Specifically, we have:

\noindent
\textbf{Step 1: \textit{Constraint Extraction}.}
  Given attributes from the grounding knowledge, extract the corresponding values from the question.
  
\noindent
\textbf{Step 2: \textit{Explicit Constraint Matching}.}
  Find values in the grounding knowledge that are correlated or coreference with the extracted constraints.
  
\noindent
\textbf{Step 3: \textit{Clarification Question Generation}.} Generate a question to clarify any misunderstanding the model couldn't resolve.

% \begin{enumerate}[leftmargin=12mm]
% \item[Step 1:]
% \textit{Constraint Extraction}.
%   Given attributes from the grounding knowledge, extract the corresponding values from the question.
% \item[Step 2:]
% \textit{Explicit Constraint Matching}.
%   Find values in the grounding knowledge that are correlated or coreference with the extracted constraints with valid values.
% \item[Step 3:]
% \textit{Clarification Question Generation}. Generate a question to clarify any misunderstanding the model couldn't resolve.
% \end{enumerate}
Note that the extracted constraint for an attribute can be a phrase, e.g., hometown: the 15th most populous city in the United States. In this case, the constraint extraction module can be seen as a question decomposer, and constraint alignment as a subquestion solver.

All steps are implemented by prompting the LLM.
For step 1, we prompt the LLM with the following instruction:
\begin{mdframed}[linewidth=1pt, linecolor=black]
\begin{quote}
  {Extract any phrases that act as conditions or constraints relating to each attribute. If you are not confident that there's an applicable phrase, signify this with `None'.
  \\
  Attributes: {\color{gray}{City, Event, Sport, Year}}
    \\
      Question: {\color{gray}{Which sport has event called America Open?}}}
\end{quote}
\end{mdframed}
To address the issue of attributes with less semantic names, such as ``Name-1'', we employ the LLM to describe the attribute using its possible values before utilizing it. The prompt is shown in Appendix D.2. This approach helps provide more context and understanding to both the model and the user.

We proceed by verifying
the value references in the grounding knowledge (Step 2). For each explicit constraint, we prompt the LLM as follows:
\begin{mdframed}[linewidth=1pt, linecolor=black]
\begin{quote}
  {Question: {\color{gray}Which sport has event called America Open?}
    \\
  For ``{{\color{gray}{America Open}}}'' in the question, identify the corresponding option it refers to. If there is ambiguity or uncertainty, where multiple options seem equally probable, or no options clearly match, respond with ``None''. 
  \\Options: {{\color{gray}{French Open, U.S. Open}}}.
  }
\end{quote}
\end{mdframed}

We collect the results as alignment feedback. 
For matches deemed successful, we categorize them as explicit clarifications, e.g., ``For `event', America Open refers to U.S. Open.''. In other cases the model returns ``None'', indicating ambiguity or uncertainty, we engage the user by posing a question to seek alignment (Step 3):
\begin{mdframed}[linewidth=1pt, linecolor=black]
\begin{quote}
     Question: {\color{gray}Which sport has event called America Open?}
    \\
    The constraint ``{{\color{gray}{America Open}}}'' in the user question is unclear.
    Ask a clarifying question to make the user confirm the corresponding value of the constraint. The constraint can refer to values in this list: {\color{gray}{French Open, U.S. Open}}
\end{quote}
\end{mdframed}

The clarifying question and the user response obtained from this interaction serve as explicit clarification, e.g., ``Question: Is the event you are referring to U.S. Open? Answer: Yes.''

\subsection{Implicit Knowledge Alignment}
At this stage, we assess the need to address implicit constraints not stated in the user's question. If needed, we identify an attribute that optimally distinguishes candidates and pose a question to assist the user in resolving any potential ambiguities or inconsistencies. 
This stage comprises three steps:

\noindent
\textbf{Step 1: \textit{Irrelevant Candidate Filtering}.} Filter out groundings that do not contribute to answering the question, using previously obtained clarifications.

\noindent
\textbf{Step 2: \textit{Distinguishable Attribute Selection}.} Identify the optimal attribute to further discern remaining candidate groundings.

\noindent
\textbf{Step 3: \textit{Clarifying Question Generation}.} If a valid attribute is identified, generate a question to ask the user about it.

% \begin{enumerate}[leftmargin=12mm]
% \item[Step 1] \textit{Irrelevant Candidate Filtering}. Filter candidate groundings with previously obtained clarifications.
% \item[Step 2] \textit{Distinguishable Attribute Selection}. Identify the optimal attribute to differentiate knowledge groundings.
% \item[Step 3] \textit{Clarifying Question Generation}. If a valid attribute is found, generate a clarifying question regarding it.
% \end{enumerate}

We begin by determining the necessity to address implicit constraints. As elaborated in Section 3, we focus on questions that pertain to a singular, distinct subject.
Given that the accurate grounding knowledge should be unique, we filter candidates and seek further clarification if multiple candidates remain.
We set aside more complex scenarios for future exploration.
Specifically, we prompt the LLM to filter the candidates (Step 1):
\begin{mdframed}[linewidth=1pt, linecolor=black]
\begin{quote}
    In the context of the given question and its clarifying information, filter the list of candidates. The aim is to select only those candidates that adhere to the conditions or constraints provided.\\Candidates:\\{\color{gray}{1. City: New York; Event: U.S. Open; Year: 2023;\\2. City: Paris; Event: French Open; Year: 2023;\\3. City: Los Angeles; Event: U.S. Open; Year: 2023;\\4. \ldots}}\\Clarifying information:\\{\color{gray}Question: Is the event you are referring to U.S. Open? Answer: Yes.}
\end{quote}
\end{mdframed}

In Step 2, we select the attribute by taking into account two aspects: (1)
Distinguishability: We aim to eliminate noisy candidates as much as possible after
clarification.
(2) Answerability: We avoid asking the user about unfamiliar attributes such as
names and ID numbers.
For simplicity, we merge Step 2 and 3 and prompt LLM with:

\begin{mdframed}[linewidth=1pt, linecolor=black]
\begin{quote}
    Given the following candidates, your task is to formulate a clarifying question to filter out irrelevant candidates. This clarifying question should aim to ascertain the value of an attribute to best differentiate among candidates. Ensure that the attribute relates to general knowledge rather than specialized knowledge.
    \\
    Candidates:\\{\color{gray}{1. City: New Yrok; Event: U.S. Open; Year: 2023;\\2. City: Los Angeles; Event: U.S. Open; Year: 2023;}}
\end{quote}
\end{mdframed}

The clarifying question and user response act as implicit clarifications, e.g., ``Question: Which city hosted America Open? Answer: L.A.''

\subsection{Answer Generation}
The final answer is generated by including explicit and implicit (if any) clarifications ($C$) in the prompt, i.e., $\text{LLM}(A|Q,K,C)$.

\subsection{A Casual Look at \textsc{MixAlign}}
\begin{figure}[htp]
  \includegraphics[width=.95\linewidth]{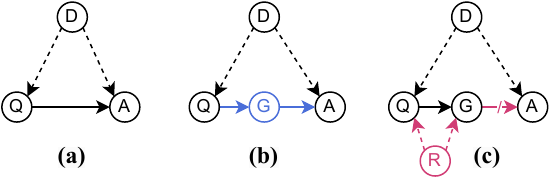}
  \centering
  \caption{Knowledge grounding ($G$) effectively boosts LLM ($D$) performance (through
    front-door adjustment) only when the knowledge is causally retrieved and can
    causally induce the answer.
    That is, the retrieval model ($R$) itself should be trustworthy enough to not
    introduce statistical co-occurrence information (i.e., a nurse must be a woman), and
    the retrieved knowledge must be aligned with the question ($Q$) in order to be utilized for further deducing the correct answer ($A$).
  }
  \label{fig:causal}
\end{figure}

To uncover the cause-effect relationships in retrieval-augmented generation, we have developed a Structural
Causal Model (SCM)~\cite{peters2017elements}. 
SCM is a directed acyclic graph that represents causal connections within a
system.

As shown in~\cref{fig:causal}(a), the pre-trained knowledge ($D$) in LLM
introduces confounding factors into the system.
For example, the model may assume that a nurse must be a woman,
resulting in biased correlations and ultimately harm model performance.
As illustrated in~\cref{fig:causal}(b), the Retrieval-augmented Language Model
mitigates biased correlations through the front-door adjustment~\cite{pearl2009causality}, which employs a
mediator ($G$, retrieved knowledge groundings) to block all directed paths from
the cause ($Q$) to the effect ($A$).
However, as depicted in~\cref{fig:causal}(c), the front-door adjustment can easily
fail when the groundings are statistically retrieved using the nearest neighbors
search based on co-occurrence information.
To address the aforementioned issue, \textsc{MixAlign} offers clear explanations on
\textit{why the question and knowledge are related}, thereby promoting front-door
adjustments and boost model performance.

%%% Local Variables:
%%% mode: latex
%%% TeX-master: "emnlp2023"
%%% End:

\section{Experiments}
% \input{tabs/dia_all}
% \subsection{Experimental Setup}
% \input{fig_dataset}
% \input{tab_template}
\label{sec:exp}
\header{Evaluation Task:}
We focus on knowledge-augmented generation instead of evidence retrieval to explore the benefits of knowledge alignment. Specifically, we consider a controlled number of irrelevant knowledge groundings (database rows) with the primary grounding in the model's input context. This count of irrelevant groundings is denoted as `Irrelevant Groundings (\#)'.

\header{Dataset: FuzzyQA} is an evolution of the OTT-QA dataset~\cite{chen2020open}. OTT-QA is an English dataset that contains open questions that require grounding on tables and text for answers. In FuzzyQA, we made two changes:

1. We shifted the focus solely to tables as the primary knowledge source (detailed in Section 3). This results in a filtered dataset comprising 1,173 (question, answer, table) triples, reserved solely for validation (our method does not require training). 

2. We simplify each question with GPT-4 by dropping constraints but ensuring the answers remain unchanged, as detailed in Appendix D.3. This adjustment was made because OTT-QA questions were crafted by annotators who had prior access to the tables, a scenario that contrasts with real-world situations where users often frame their queries without detailed table knowledge. By simplifying the questions, we aim to simulate this real-world ambiguity. Note that this simplification makes the questions more challenging for LLMs, as the reduced detail introduces extra complexity.

\begin{table}[t]
\small
\centering

\label{tab: info}
\begin{tabularx}{\linewidth}{p{1.2cm}X}
\toprule
Type & Clarifying Information\\
\midrule 
Semantic & The term `A' in the question refers to `B' in our database.\\
\midrule
Contextual &  The value for the missing contextual condition in the question is `A'.\\
\midrule
Structural & The value for the condition `B' in the database is `A'.\\
\midrule
Logical & The complex condition `A' in your question refers to the condition `B' in the database.\\
\bottomrule
\end{tabularx}
% }
\caption{Oracle clarifying information for each knowledge alignment type. A and B denote human and domain knowledge, respectively.  
}
\end{table}

\header{Language Model and Baselines.}
\textsc{MixAlign} is designed to be compatible with any LLM, in this section, we employ the OpenAI Text-DaVinci-003
(176B)~\cite{ouyang2022training} for all the methods.
We examine the impact of incorporating clarifications into the prompt:

\bullethdr{None}. No clarification 
 included.

\bullethdr{Oracle}. We reverse-generate the oracle clarifications from the ground-truth answer and knowledge considering the templates in~\cref{tab: info} with GPT-4. The prompt is shown in Appendix D.1.
% The clarifying information is generated considering the template as shown in~\cref{tab: info}.

\bullethdr{Direct-Ask}~\cite{kuhn2022clam}. Clarifying questions are asked based solely on the original question. Direct-Ask prepends the question with a prompt: ``{In order to answer this question, I have to ask the following clarifying question:}''.
% We modified the origin prompt to prevent repeating the user's question back to them.

\bullethdr{Knowledge-Ask}. Building upon Direct-Ask, we incorporate candidate knowledge to generate clarifying questions and modify the instruction as ``{In order to answer this question with the context, candidate 1, candidate 2, \ldots, I have to ask the following clarifying question:}''.

\bullethdr{\textsc{MixAlign} (proposed)}. In addition to the previous settings, we introduce alignment feedback to enhance the process of posing clarifying questions.

\begin{figure*}[htp]
\centering  
\includegraphics[width=\linewidth]{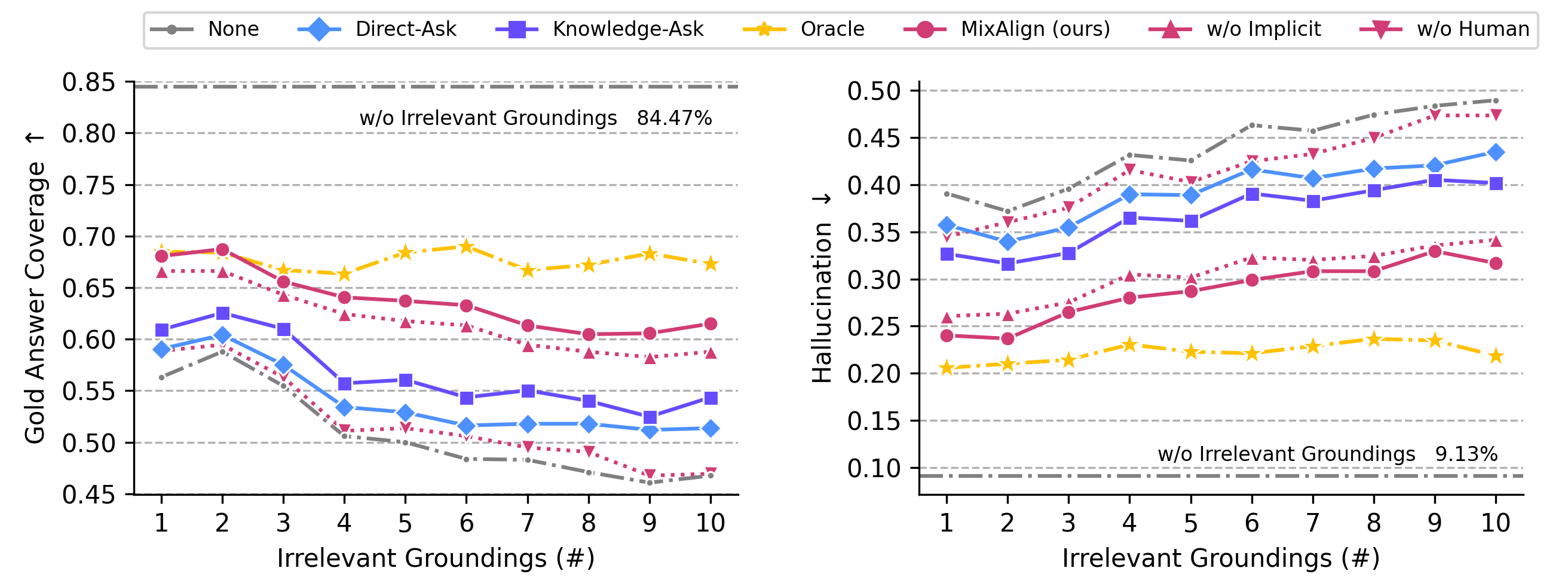}
\caption{Overall evaluation results. We fused irrelevant knowledge groundings with ground-truth evidence for LLM generation. ``w/o Implicit'' and ``w/o Human'' represent two ablation variants of \textsc{MixAlign} (detailed in Sec. 5.2.) "w/o Irrelevant Groundings" denotes that the LLM is solely reliant on the ground-truth evidence, absent of noise. We report the mean from one run over the FuzzyQA dataset. OTT-QA results are in Appendix A.1.
}\label{fig:overall_eval}
\end{figure*}
\header{Metrics.}
We adopt G-EVAL~\cite{liu2023gpteval}, a state-of-the-art ChatGPT-based evaluation framework, and consider the three metrics below. Please refer to Appendix B for details.

\bullethdr{Gold Answer Coverage}. This binary metric evaluates whether the model's answer covers the gold answer, indicating how accurately the model captures the relevant information.

\bullethdr{Hallucination}. This binary indicator detects factual inconsistencies between the model's answer and the input context, highlighting instances where the model generates unsupported information.

\bullethdr{Accepted}. This binary indicator checks whether the clarifying question posed by the model (partially) repeats the original question back to the user.

\header{User Simulator.}
Following~\citet{kuhn2022clam}, we implement the user simulator as an ``oracle''
language model that has access to attributions about the gold answer subject (detailed in Appendix D.4).
% This information is incorporated into the prompt to ensure the provision of a reliable and clarifying answer.

\subsection{Overall Results}
\textbf{Negative impact of irrelevant groundings.}
Comparing ``None'' to ``w/o irrelevant groundings'' in~\cref{fig:overall_eval} highlights the negative effect of including irrelevant groundings. Specifically, 
coverage decreases by $25.7\%$ to $39.4\%$, and hallucination increases by $28.1\%$ to $39.8\%$.
Moreover, as more irrelevant groundings are introduced, performance worsens because they become increasingly difficult for the LLM to differentiate and associate.

\textbf{Knowledge alignment significantly boosts the LLM performance.} With oracle clarification, we observe a noticeable gap of $9.6\%$ to $22.2\%$ in coverage and $16.2\%$ to $27.1\%$ in hallucination.
While the results are promising, a significant gap remains when compared to  ``w/o irrelevant groundings''.
One key reason for this is the complexity of the prompt given to the LLM. In the absence of irrelevant groundings, the LLM's prompt contains only a single ground-truth grounding. However, with clarification, the prompt is populated with multiple clarifying details and all available groundings. This increased complexity poses an inherent challenge for the LLM to process effectively.

\begin{table}[tp]
\small
\centering

\label{tab: accept}
\begin{tabular}{ccc}
\toprule
\textbf{Direct-Ask} & \textbf{Knowledge-Ask}  & \textsc{MixAlign}  \\
\midrule
$36.09 \pm 0.95$ & $27.67 \pm 0.52$ & $100$ \\
\bottomrule
\end{tabular}
\caption{Average acceptance rate (\%) for clarifying questions across varied irrelevant groundings (\#).}
\end{table}

\textbf{\textsc{MixAlign} outperforms baseline clarifying methods.}
When benchmarking \textsc{MixAlign} against Direct-Ask, we note a difference of $8.1\%$ to $11.7\%$ in coverage and $9\%$ to $11.8\%$ in hallucination. In comparison with Knowledge-Ask, the disparity is $4.6\%$ to $8.9\%$ in coverage and $6.3\%$ to $9.2\%$ in hallucination. The key driver behind this enhancement is \textsc{MixAlign}'s feedback-driven clarifying question generation. This approach ensures \textsc{MixAlign} clarifies unaligned content instead of seeking generic clarifications and echoing (in part or whole) the initial question back to the user, which is typically unhelpful and unanswerable.
This is further illustrated in~\cref{tab: accept}, where the acceptance rate for baseline methods lags significantly behind \textsc{MixAlign}. Specifically, Direct-Ask tends to mirror the original question, while Knowledge-Ask tends to
focus on the compositional gap of the original query and pose
the subquestion back to the user. See Appendix A.4 for a detailed error analysis.

\begin{table*}[htp]
\small
\centering

% Method & Case Clarifying Question\\
% \midrule 

\label{tab: case}
\begin{tabularx}{\linewidth}{p{2.5cm}XX}
\toprule
\textbf{User question} & What are the campaign notes of the 2012 Democratic candidate born on May 23, 1958? & Where did the Magic: The Gathering player known as ``The Great One'' win? \\
\midrule
\midrule
\textbf{Direct-Ask} & What specific information are you looking for regarding the 2012 Democratic candidate? & What format of Magic: The Gathering was ``The Great One'' playing? \\
\midrule
\textbf{Knowledge-Ask} & Which candidate was born on May 23, 1958? & Which season did \textbf{Jon Finkel} win? \\
\midrule
\textsc{MixAlign} (ours) & Is the candidate you are referring to Keith Judd? & Which season did ``The Great One'' win? \\ 
\bottomrule
\end{tabularx}
% }
\caption{Case clarifying questions. Baseline methods may: 1. inquire about irrelevant factors that do not aid problem-solving, 2. reflect the question back to the user, and 3. incorporate noisy knowledge into the question.
}
\end{table*}

Remarkably, the performance of \textsc{MixAlign} is nearly on par with that of oracle clarification when the number of irrelevant groundings is minimal, as a limited number of candidates makes it easier for the LLM to achieve precise constraint matching. But with more candidates, distinguishing between them becomes more challenging, causing the performance gap to widen to as much as $7.7\%$ in coverage and $9.8\%$ in hallucination.
% We will further investigate this in the following ablation study.

\begin{figure}[t]
\centering  
\includegraphics[width=.9\linewidth]{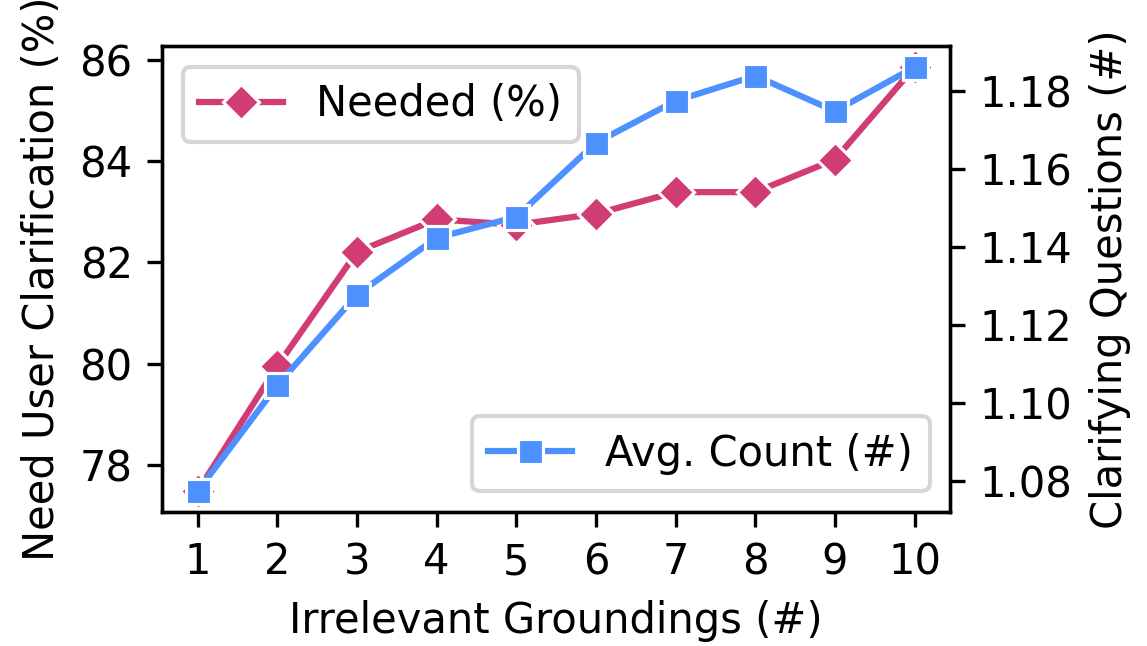}
\caption{Efficiency analysis results of \textsc{MixAlign}. We target:
1. Needed (\%):  How often do we need to request user clarification after the initial LLM-based constraint matching?
2. Avg.Count (\#): When needed, how many clarifying questions, on average, are posed?}
\label{fig:efficiency}
\end{figure}

\subsection{Ablation Study on \textsc{MixAlign}}
We consider two ablation variants of \textsc{MixAlign}: 1. w/o Implicit: This version excludes implicit knowledge alignment and only addresses constraints that are explicitly stated in the user's question.
2. w/o Human: This version relies solely on LLM-based constraint matching and does not incorporate additional human-assisted clarification.
The results are merged into~\cref{fig:overall_eval}.
Our key findings are:

\textbf{Human assistance is exceptionally vital.} 
For the w/o Human version, we note a significant reduction in performance when solely depending on LLM-based constraint matching. As the number of irrelevant groundings increases, this reduction becomes more noticeable. This suggests that the LLM struggles to extract and match constraints with high confidence. Further evidence of this is seen in the increasing difference between the w/o Human and w/o Implicit versions, emphasizing that more constraints are not confidently matched, leading to the need for human clarification.

\textbf{Implicit knowledge alignment is necessary.}
We see that removing implicit knowledge alignment consistently leads to reduced performance.
Comparing w/o Implicit to \textsc{MixAlign}, we also observe that the performance gap remains largely constant, irrespective of the increase in irrelevant groundings.
This is attributed to error propagation from the explicit knowledge alignment, as implicit alignment targets only the groundings remaining after explicit alignment. Future work could consider an end-to-end approach to mitigate this limitation.

\subsection{Efficiency Analysis of \textsc{MixAlign}}
In this section, we assess the user effort required by \textsc{MixAlign} for clarifying misalignments.
As shown in~\cref{fig:efficiency}, as the number of irrelevant groundings increases, there's a rise in both the percentage of examples requiring user input and the average number of questions asked.
However, \textsc{MixAlign} successfully reduces the need for user clarifications by $14.2$ to $22.5\%$. Furthermore, with an average question count spanning from $1.08$ to $1.19$, showing that usually just one single clarifying question is needed, which verifies the efficiency.
See Appendix A.3 for a detailed cost analysis.

%%% Local Variables:
%%% mode: latex
%%% TeX-master: "emnlp2023"
%%% End:

\section{Conclusion and Future Work}
In this work, we introduce the \textit{Knowledge Alignment} problem, which addresses mismatches between constraints present in user questions and the knowledge groundings referred to by LLMs,
% This ensures that the model can more effectively utilize these groundings for enhanced and faithful generation.
% Facing this challenge, 
and we propose the \textsc{MixAlign} framework to bridge this gap by generating clarifications for any identified misalignments.
Experimental results highlight the crucial role of knowledge alignment in improving model performance and faithfulness and demonstrate the efficacy of \textsc{MixAlign} in generating high-quality clarifications. 
Future studies could explore adapting \textsc{MixAlign} to various knowledge modalities, thereby broadening its applicability.
\section*{Limitations}
% In this work, we utilize the OpenAI Text-DaVinci-003 model as the LLM backbone for our evaluations. While this offers a robust foundation, future work might consider integrating diverse models to achieve a more comprehensive assessment.

MixAlign reduces, but does not eliminate, the occurrence of hallucinations. By introducing explicit clarifications, we build a causal link between the question and the evidence that aids the LLM in more accurately deducing answers as it creates a clearer pathway for reasoning. However, since our method does not establish a definitive causal path for deriving answers from the question and evidence, hallucinations can still occur, emphasizing the need for future research.

External knowledge extends beyond simple tabular databases or textual formats, often manifesting with more complex modularity which combines different elements together. Addressing these intricate configurations poses a formidable challenge, and we outline this as an area for future exploration.

A further limitation to consider is the increased computational load and time consumption associated with the additional clarification steps. We mitigate this in our study by involving model-based alignment and avoiding multi-turn dialogues for human-assisted alignment. Nevertheless, more efficient strategies for addressing this concern could be further investigated.
\section*{Acknowledgements}
Shuo Zhang and Junzhou Zhao were supported in part by National
Key Research and Development Plan in China
(2023YFC3306100) and National Natural Science Foundation of China (62272372, 61902305,
U22B2019). Liangming Pan and William Yang Wang were not supported by any of the above projects.

% Entries for the entire Anthology, followed by custom entries
\bibliography{anthology,custom}
\bibstyle{acl_natbib}
\appendix

\section{Additional Experimental Results}
\subsection{Results on the OTT-QA dataset}

\begin{figure*}[htp]
\includegraphics[width=\linewidth]{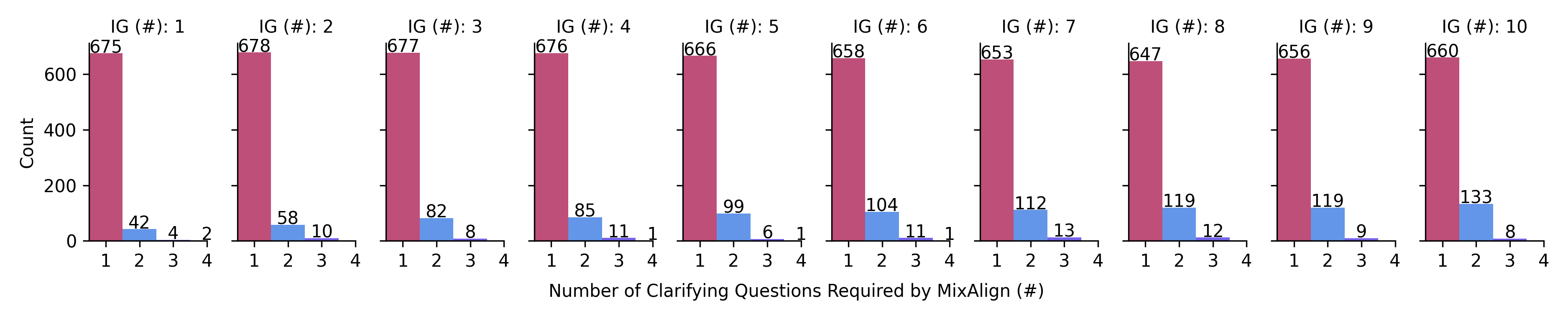}
  \centering
  \caption{Distribution of clarifying questions based on the count of irrelevant groundings, denoted as IG(\#).}
  \label{fig:full_num}
\end{figure*}

\begin{figure}[t]
\includegraphics[width=\linewidth]{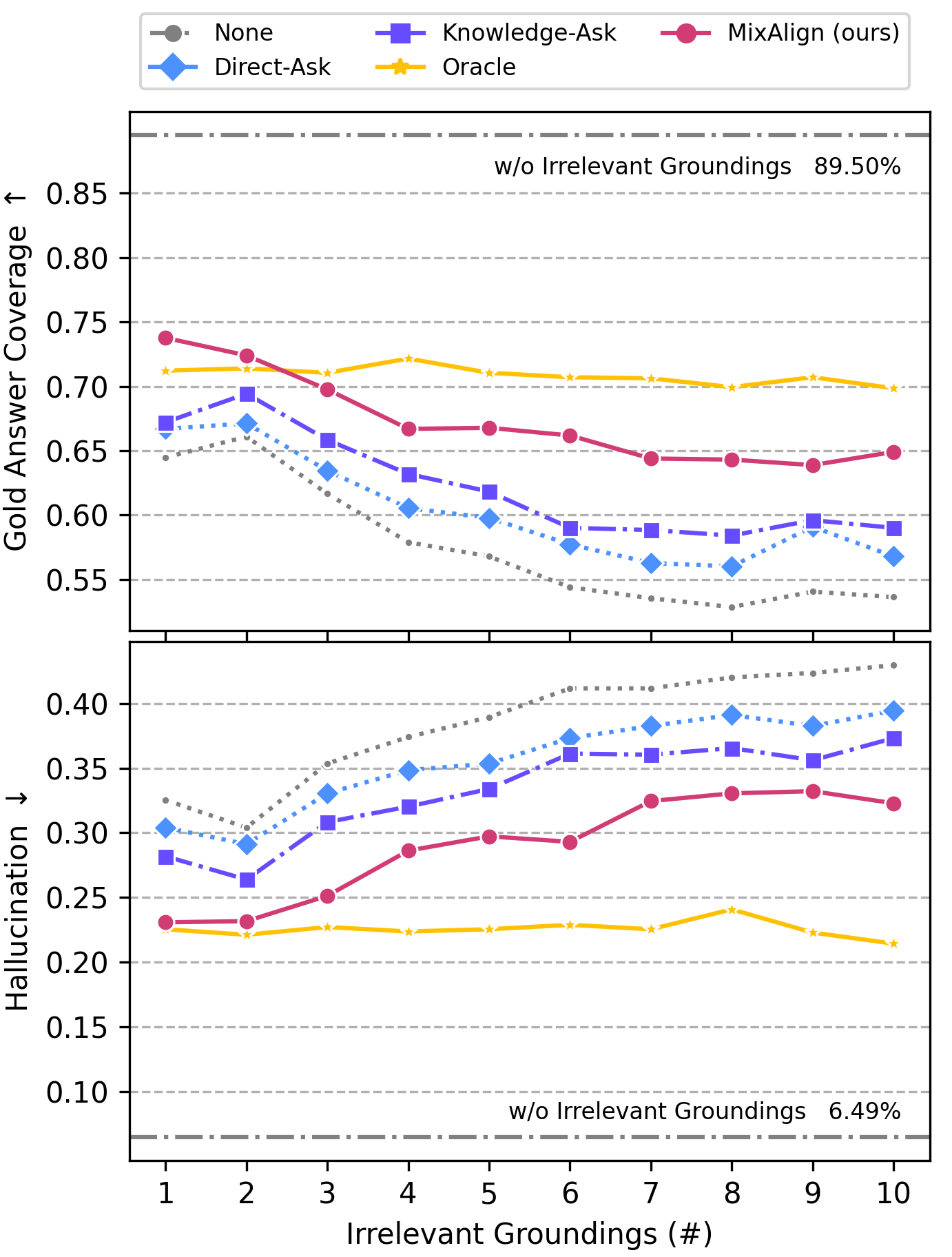}
  \centering
  \caption{Overall results on OTT-QA. We report the mean over the 1,173 instances corresponding to FuzzyQA examples.}
  \label{fig:ottqa}
\end{figure}

To study the impact of question simplification on performance, we further conduct experiments on the original OTT-QA questions. As depicted in Figure~\ref{fig:ottqa}, the findings in Section 5.1 still hold, indicating that question simplification isn't the primary contributor to the observed performance discrepancy. We noticed an improvement of $5\%$ in terms of coverage and hallucination relative to the FuzzyQA results shown in Figure~\ref{fig:overall_eval}. This supports the aforementioned explaination that question simplifications indeed pose a greater challenge for LLMs. Specifically, this question simplification particularly influences contextual knowledge misalignment where users may omit constraints.

Surprisingly, MixAlign outperforms the oracle in terms of coverage when the number of irrelevant groundings is minimal. This phenomenon might be attributed to our decision to reuse oracle clarifications from FuzzyQA in this evaluation. While we assumed that the clarifications in OTT-QA would fall within those from FuzzyQA, the results hint at possible inconsistencies in the oracle data.

\subsection{Distribution of Clarifying Questions in MixAlign}
In this section, we illustrate the comprehensive distribution of clarifying questions necessitated by MixAlign, as depicted in \cref{fig:full_num}. We see that the maximum number of clarifying questions posed is $4$. With an increase in irrelevant groundings, there is a corresponding increment in the number of clarifying questions. Nevertheless, the distribution predominantly hovers around a single question. It is noteworthy that the surge is primarily in instances requiring two questions; however, occurrences necessitating three or four questions do not exhibit a substantial increase.

\subsection{Computational Efficiency and User Experience Analysis of MixAlign}
\textbf{Computational Costs.}
We evaluate the computational cost with the \textit{Number of LLM Inferences Required per Forward Pass}.
For a forward pass in MixAlign, key operations include constraint extraction (plus attribution description, if applied), Explicit Constraint Matching, and Clarification Question Generation that is undertaken if the Explicit Constraint Matching or Irrelevant Candidate Filtering does not yield confident results. Therefore, the total number of inferences ranges from $const + 2$ to $2\times const + 3$, where $const$ is the number of explicit constraints extracted. Our results with Text-Davinci-003 reveal this resource cost is in an average of $3.7$ inferences per example in our dataset.
Regarding time cost, we observed an extra latency of 4.87 seconds brought by MixAlign. Due to the variability introduced by factors like GPU performance and API latency, we choose to consider it proportional to the number of inferences rather than pursuing detailed quantification.

\textbf{User Experience.}
The trade-off between model accuracy and the additional effort required from users to answer clarification questions underpins the user experience with MixAlign. This balance varies across applications; in contexts where accuracy is paramount, the inclination to ask additional questions is less of a concern. An additional observation concerns the nature of some questions generated by MixAlign, which may require users to select one option from several, potentially increasing the burden on the user. We will include the discussions in our final manuscript.
Our efforts in this work aim to minimize the need for human clarifications. As shown in Figure 6 and Appendix A.2, our design ensures that on average 14-22\% of instances do not require human intervention. 

\subsection{Error Analysis of MixAlign}
A detailed error analysis is crucial for understanding the scenarios in which MixAlign might fail or underperform. To this end, we investigate the question: ``What makes a hard example for MixAlign?''

As shown in Figure~\ref{fig:overall_eval} (w/o Human), without human clarifications, automatic knowledge alignment with constraint extraction and matching offers limited performance improvements. Generally, the extraction process is reliable except in cases involving complex logical conditions casual expressions in the question or complex and vague attributes, like ``electorate'' or ``ethnic group S estimate'' in the retrieved knowledge groundings. The main source of errors, however, lies in constraint matching. MixAlign seeks human clarifications when the LLM struggles to confidently link constraints. Achieving this level of confidence is challenging, even for advanced models like text-davinci-003 (176B). In summary, complex or ambiguous attribute and value names in the retrieved knowledge, and complex or causal user questions (as the extracted constraint, even correct, is hard to link), can lead to challenging scenarios for MixAlign.

To further understand the factors that influence MixAlign's performance, we conducted a quantitative analysis of attribute and value numbers on our dataset that spans attribute counts from $3$ to $7$ with distributions of $185$, $442$, $334$, $210$, and $3$.
\begin{table}[h]
    \centering
    \caption{Coverage and Hallucination Percentages by Attribute Count}
    \label{tab:attributes}
    \resizebox{0.47\textwidth}{!}{%
    \begin{tabular}{@{}cccccc@{}}
        \toprule
        \textbf{\#Attributes} & \textbf{3} & \textbf{4} & \textbf{5} & \textbf{6} & \textbf{7} \\ \midrule
        \textbf{Coverage\%}   & 68.38      & 66.18      & 68.65      & 66.14      & 100        \\
        \textbf{Hallucination\%} & 27.41    & 28.82      & 28.29      & 32.33      & 0          \\
        \bottomrule
    \end{tabular}%
    }
\end{table}

As shown in Table~\ref {tab:attributes}, an increase in the number of attributes does not necessarily lead to a decrease in performance (we ignore the \#Attr=7 results as there are only 3 examples.) Note that the increased context length due to the increased attributes can also lead to performance drop.
Besides the number of attributes, the number of values under a certain attribute, proportional to \#irrelevant groundings, does affect constraint matching efficiency due to the increased complexity of the co-reference task. As shown in Figure~\ref{fig:overall_eval}, this results in a 5\% drop in coverage and an 8\% increase in hallucination

We further conducted additional experiments to analyze MixAlign's performance across various types of misalignments.

\begin{table}[h]
    \centering
    \caption{Coverage and Hallucination Percentages by Misalignment Type and Count}
    \label{tab:misalignment_stats}
    \resizebox{.47\textwidth}{!}{%
    \begin{tabular}{@{}lcc@{}}
        \toprule
        \textbf{Misalignment Type}   & \textbf{Coverage\%} & \textbf{Hallucination\%} \\ \midrule
        Contextual Misalignment      & 64.57               & 33.00                    \\
        Logical Misalignment         & 72.50               & 26.15                    \\
        Semantic Misalignment        & 67.07               & 30.12                    \\
        Structural Misalignment      & 62.22               & 33.57                    \\ \midrule
        \multicolumn{3}{c}{\textbf{Overall Misalignment Counts}}                       \\ \midrule
        Single                       & 70.47               & 25.28                    \\
        Multiple                     & 68.02               & 31.54                    \\
        \bottomrule
    \end{tabular}%
    }
\end{table}

From Table~\ref{tab:misalignment_stats}, we find that MixAlign is more effective in addressing logical and semantic misalignments. Moreover, encountering multiple misalignments simultaneously leads to a decrease in MixAlign's performance, though not severe.

\section{Details on Evaluation}
Please note that GPT-4 refers to version gpt-4-0613, which includes data up until Sep. 2021.

\subsection{Coverage and Hallucination}
We follow the G-EVAL framework and employ OpenAI GPT-4 for evaluation. To enhance efficiency, we also implement rule-based matching to determine whether the gold standard answer is contained within the generated response prior to the LLM-based evaluation. 
For gold answer coverage, we prompt the LLM with:
\begin{mdframed}[linewidth=1pt, linecolor=black]
\begin{quote}
Human Evaluation of Question Answering Systems:
\\Coverage: Is the model answer consistent with the gold answer?\\Gold Answer:{\color{gray}Golf}\\Model answer: {\color{gray}The sport you are asking for is golf.}\\Does the model answer cover the gold answer?
\end{quote}
\end{mdframed}

For hallucination, we follow the prompt in G-EVAL:
\begin{mdframed}[linewidth=1pt, linecolor=black]
\begin{quote}
Human Evaluation of Question Answering Systems:
\\Factual Consistency: Does the system answer untruthful or misleading facts that are not supported by the question, context, and gold answer?\\Question:{\color{gray}Which sport has an event called America Open?}\\Context:{\color{gray}City: New York; Event: U.S. Open;
Year: 2023; Sport: Tennis\\City: Los Angeles; Event: U.S. Open;
Year: 2023; Sport: Golf.}\\Gold Answer: {\color{gray}Golf}\\Model answer: {\color{gray}The sport you are asking for is golf.} \\Does the model answer contain factual inconsistency?
\end{quote}
\end{mdframed}

\subsection{Accepted}
A clarifying question is considered unacceptable if it either 1. echoes the original user question or a subquestion back to the user, or 2. elicits negative user responses such as ``I don't know.'' 

We identify these unsatisfactory outcomes by examining the user response. Specifically, we have compiled a list of potential answers to user questions and subquestions, along with expressions of uncertainty or lack of knowledge, such as ``I don't know'' or ``sorry.'' Through rule-based matching, we assess whether a user's answer contains these phrases; if so, we categorize the corresponding clarifying question as unacceptable.

\section{Forward View on Adapting MixAlign to Various Knowledge Modalities}
MixAlign leverages the LLM to match constraints in the question and constraints in pieces of evidence, identifying mismatches for further clarification. To adapt MixAlign for different knowledge formats, a straightforward approach would be using specialized information extraction (IE) models to transform evidence constraints into textual forms for LLM comparison. However, standard IE tools could result in errors that affect the LLM's performance in our MixAlign framework. A more effective approach might involve employing a stronger LLM, such as GPT-4, for the IE process, thereby providing higher quality input to MixAlign, i.e., strong-to-weak generalization.

This leads us to an intriguing research question: \textit{Is it feasible to preprocess inputs using a weaker LLM for a stronger LLM?} This concept entails equipping a ``strong brain'' with a ``less accurate eye,'' a notion not extensively explored in current literature. While there is existing research on training models from a weak-to-strong generalization perspective~\cite{collinws}, the specific application of this approach at the input level remains uncharted territory. Future work might include modifying the format of IE outputs, transitioning from basic labels to more descriptive formats such as textual explanations or probability logits. 
Such modifications could optimize the utilization of weaker models within the overarching process, leveraging their strengths more effectively.

\section{Supplementary Details on Prompts}
GPT-3 denotes Text-DaVinci-003. GPT-4 denotes the 0613 version. Here, we detail the prompts used for oracle clarification, dataset processing, and simulated interaction.

\quad

\subsection{Oracle Clarification Generation (GPT-4)}
\begin{mdframed}[linewidth=1pt, linecolor=black]
\begin{quote}
Given the question and its gold database knowledge detect if there exist misalignments regarding each type below.\\\\Semantic Misalignment: The user might use an ambiguous term in the question that, while ideally should map to a single item in the database, in reality can correspond to multiple columns or values, leading to uncertainty about the specific item the term refers to. For instance, if a user asks ``What is the best burger?'', the term ``best burger'' could refer to different columns such as taste and price. In another case, if a user mentions ``Paris'', it's ambiguous whether it refers to ``Paris, France'' or ``Paris, Texas''.
\\\\Contextual Misalignment: The user may omit certain conditions in the question, making it hard to relate with the knowledge. For example, for the question ``What is the 15th most populous city in the United States?'', the statistics might change over time, so without specifying the year in the question, there's a misalignment with the context.
\\\\Structural Misalignment: The user could state conditions that are not covered in the database structure. For example, if a user asks ``Find me an American writer'', this question cannot be answered if the database does not include nationality information for writers.
\\\\Logical Misalignment: The user's question might contain intricate conditions where certain aspects need to be resolved before others can be clarified. This often occurs when a single condition in the question encapsulates other sub-conditions or questions that need to be addressed first. For example, in the query ``Find me a movie directed by the singer who has won a Grammy'', the identification of the Grammy-winning singer is a prerequisite before the movies directed by this person can be determined.
\\Question: {\color{gray}Which sport has an event called America Open?}\\Knowledge: {\color{gray}City: Los Angeles; Event: U.S. Open; Year: 2023;}",
\end{quote}
\end{mdframed}

\subsection{Attribute Description (GPT-3)}
% Note: GPT-3 denotes Text-Davinci-003.
\begin{mdframed}
[linewidth=1pt, linecolor=black]
\begin{quote}
{Column names along with potential values:\\{\color{gray}Note: 2021, 2022, 2023\\
City: New York, Los Angeles, Paris}\\Generate a concise phrase that accurately describes each column name. If the column names lack sufficient semantic clarity or descriptiveness, furnish them with additional context or explanations.}
\end{quote}
\end{mdframed}

\subsection{Question Simplification (GPT-4)}
\begin{mdframed}[linewidth=1pt, linecolor=black]
\begin{quote}
{Question: {\color{gray}Which 2010 Regional League Division 2 Southern Region team plays at the stadium with the largest capacity?}\\
Simplify this question by dropping attributes and conditions such as time and place, make sure that the answer to the simplified question is the same as the original question.}
\end{quote}
\end{mdframed}

\subsection{User Simulator (GPT-4)}
\begin{mdframed}[linewidth=1pt, linecolor=black]
\begin{quote}
You are a user of a QA system. You know: {\color{gray}City: Los Angeles; Event: U.S. Open;
Year: 2023;}. 
\\You just asked '{\color{gray}Which sport has an event called America Open?}' and the system throws back a clarifying question '{\color{gray}Is the America Open referring to U.S. Open?}'.
\\
Please answer the clarifying question precisely. Do not respond with anything else besides the primary subject asked by the clarifying question.
\end{quote}
\end{mdframed}

\section{Licensing and Terms for Datasets}
In this study, we developed the FuzzyQA dataset, based on OTT-QA. Like OTT-QA, FuzzyQA adheres to the MIT License, reflecting our commitment to legal compliance and respecting OTT-QA's original terms. This licensing approach ensures transparency and aligns with legal and ethical usage standards. Our enhancements to OTT-QA for FuzzyQA align with the original dataset's intended research and academic applications

% \section{Example Appendix}
% \label{sec:appendix}

% This is a section in the appendix.

\end{document}